\documentclass[letterpaper, 10 pt, conference]{ieeeconf}  

\IEEEoverridecommandlockouts                              

\usepackage{amsmath,amssymb,amsfonts}
\usepackage{algorithm}
\usepackage{algorithmic}
\usepackage{graphicx}
\usepackage{textcomp}
\usepackage{xcolor}

\usepackage{svg}

\usepackage{tikz}
\usetikzlibrary{shapes.geometric, arrows}
\def\BibTeX{{\rm B\kern-.05em{\sc i\kern-.025em b}\kern-.08em
    T\kern-.1667em\lower.7ex\hbox{E}\kern-.125emX}}

\title{Malicious Path Manipulations via Exploitation of Representation Vulnerabilities of Vision-Language Navigation Systems\\
}

\author{Chashi Mahiul Islam\textsuperscript{*}, Shaeke Salman\textsuperscript{*}, Montasir Shams, Xiuwen Liu, and Piyush Kumar 
\thanks{*Denotes equal contributions}
\thanks{ All authors are with the Department of Computer
Science, Florida State University, Tallahassee, Florida 32306, USA. Email:
{\tt\small
ci20l@fsu.edu, salman@cs.fsu.edu, mshams@fsu.edu, liux@cs.fsu.edu, piyush@cs.fsu.edu}}%
}

\begin{document}

\maketitle

\begin{abstract}
Building on the unprecedented capabilities of large language models for command understanding and zero-shot recognition of multi-modal vision-language transformers, visual language navigation (VLN) has emerged as an effective way to address multiple fundamental challenges toward a natural language interface to robot navigation. However, such vision-language models are inherently vulnerable due to the lack of semantic meaning of the underlying embedding space. Using a recently developed gradient-based optimization procedure, we demonstrate that images can be modified imperceptibly to match the representation of totally different images and unrelated texts for a vision-language model. Building on this, we develop algorithms that can adversarially modify a minimal number of images so that the robot will follow a route of choice for commands that require a number of landmarks. We demonstrate that experimentally using a recently proposed VLN system; for a given navigation command, a robot can be made to follow drastically different routes. We also develop an efficient algorithm to detect such malicious modifications reliably based on the fact that the adversarially modified images have much higher sensitivity to added Gaussian noise than the original images.
\end{abstract}

\section{Introduction}

In recent years, the landscape of artificial intelligence has been significantly reshaped by the advent of large language models (LLMs). These models have demonstrated unparalleled proficiency across a wide range of natural language processing tasks, exhibiting a remarkable ability to generate human-like text. Simultaneously, the advent of multimodal models that combine LLMs with visual processing capabilities has introduced new possibilities for AI applications, offering unprecedented capabilities in zero-shot classification. The development of Vision-and-Language Navigation (VLN) systems (\cite{anderson2018evaluation,chen2020touchdown}) exemplifies this trend, offering new capabilities such as guiding robots or assisting users in virtual environments based on combined visual and textual instructions or queries. For example, GPT-4V has been used to generate detailed instructions for an autonomous driving system under many real-world scenarios~\cite{wen2023road}. LM-Nav~\cite{shah2022lmnav}, a recently developed VLN system, enables a real-world robot to navigate through complex outdoor environments via natural language commands without any customized training of the models. 

However, despite these advancements, the vulnerabilities of LLMs and multimodal models to adversarial attacks pose significant challenges. Adversarial attacks, which involve subtly altered inputs designed to deceive AI models, highlight the importance of robustness in these systems (\cite{goodfellow2015explaining,szegedy2014intriguing}). Studies have indicated that vision transformers may exhibit more resilience to certain types of adversarial attacks than other architectures (\cite{bhojanapalli2021understanding,qin2023understanding}); however, the semantic meaning and the representations produced by these models, especially in the context of adversarial resilience, remain less explored areas. Our research, as detailed in (\cite{salman2024intriguing,salman2024zshot,salman2024unalign}), has shown that the representations by transformers might have limited semantic meaning, underscoring the need for further investigation into the vulnerabilities and implications of these representations.

This paper delves into the vulnerabilities associated with the representations produced by vision-language navigation systems and multimodal models. By designing algorithms that can manipulate and alter routes without noticeable changes to the image labels or content, we demonstrate the potential for adversarial exploitation. To the best of our knowledge, this is the first work on adversarial exploitation of vision language navigation and this exploration not only sheds light on the robustness of current VLN models but also paves the way for future research to enhance the security and reliability of AI systems operating in multimodal domains.

\section{Related Work}

Vision-and-Language Navigation (\cite{anderson2018evaluation,chen2020touchdown}) is an inter-disciplinary research area that requires the integration of both visual perception and natural language understanding. In VLN, agents are tasked with navigating through unfamiliar environments (both persistent and non-persistent) based on natural language instructions (\cite{anderson2018evaluation,chen2020touchdown,anderson2018visionandlanguage,krantz2023iterative}). Several datasets have been curated specifically for VLN tasks, with the Room-to-Room (R2R) dataset emerging as a leading benchmark \cite{anderson2018visionandlanguage}. Over recent years, VLN has attracted considerable attention due to its potential applications in autonomous navigation, robotics, and virtual reality. Robotic VLN enables robots to navigate and map environments using visual perception and natural language instructions. Several approaches have been proposed to address this problem (\cite{anderson2018visionandlanguage,li2022unieden}). Most of the approaches require pre-training and fine-tuning models. For instance, Majumder \textit{et al}. \cite{majumdar2020improving} pretrain a VLN-BERT model using extensive image-text pairs sourced from the web. In a separate study, Chen \textit{et al}. \cite{10.1145/3526024} incorporate directional cues into the conventional sequence-to-sequence VLN framework to augment performance. Additionally, some methods rely on visual language and memory to guide robot navigation by selecting landmarks and using prompt guidance \cite{10025736}.

The recent development of the Vision-Language Model (VLM) introduces the concept of zero-shot learning, implying that pretrained VLMs can predict outcomes for tasks they have not seen earlier. For example, the CLIP model, recently introduced by Radford \textit{et al}. \cite{radford2021learning}, utilizes image-text contrastive objectives to achieve such capabilities. Pretrained multimodal models like CLIP and GPT-3 \cite{brown2020language} have been used in the field of robotic navigation \cite{shah2022lmnav}. These models leverage their capabilities to make navigational decisions by mapping their implicit knowledge about the semantic context of the environment into inputs for robot motion planning \cite{dorbala2023embodied}. They have been trained on large and diverse datasets with weak supervision, allowing them to produce generalizable solutions for navigation tasks \cite{shah2023vint}. These models have been shown to outperform specialist models trained on singular datasets and exhibit positive transfer to a variety of downstream navigational tasks (\cite{krantz2023iterative,yang2023transferring}). LM-Nav, developed by Shah \textit{et al}. \cite{shah2022lmnav}, facilitates embodied instruction following without requiring fine-tuning. It leverages large-scale pretraining and off-the-shelf models to achieve this capability. The model showcases its ability to execute natural language instructions in complex environments, even in the absence of human-annotated navigation data, underscoring its capability for generalization across considerable distances.

While large multimodal models demonstrate impressive performance and robustness in standard environments, these deep learning models are susceptible to deception or disruption when confronted with carefully crafted adversarial inputs. Even imperceptible changes to the input can prompt these models to change their predictions. Most of the adversarial attacks exploit particular characteristics of the classifier. Recently, Wen \textit{et al}. \cite{wen2024secure} have shown a Navigational Prompt Suffix (NPS) attack targeted to LLM-based navigation models. Several recent studies have also addressed adversarial attacks on multimodal models, which can potentially jailbreak aligned LLMs or VLMs (\cite{carlini2023aligned,zou2023universal,shi2023large,qi2023visual}). Our prior works (\cite{salman2024intriguing, salman2024unalign}) are the first ones focusing on representation vulnerabilities of vision-language models. In this paper, we investigate how representation vulnerabilities in vision-language models can be exploited to modify routes in VLN systems. We demonstrate the feasibility of these exploits by implementing them in LM-Nav~\cite{shah2022lmnav}. To the best of our knowledge, this is the first study on exploiting
the representation vulnerabilities in the VLN systems. Furthermore, we also develop an efficient and reliable method to detect adversarial modifications based on their sensitivity to added Gaussian noise to modified images rather than unmodified natural images.

\section{Preliminaries}

Vision-and-Language Navigation systems (\cite{shah2022lmnav, anderson2019chasing,chen2020topological, huang2023visual}) are composed of a few key components that interface between language, vision, and robot control. Initially, a language model parses instructions into salient landmarks. Next, a vision-language model grounds these landmarks in the visual observations. Finally, a visual navigation model controls the robot between specified waypoints.

In this work, we present the representation vulnerabilities of the VLM in vision-language navigation. To demonstrate the effectiveness of our method, we use the LM-Nav system, which consists of all three previously mentioned components: \textbf{LLM} for instruction to landmark conversion, \textbf{VLM} for grounding images with landmarks, and \textbf{VNM} for navigating robots.

\subsection{Core Components}

LM-Nav consists of three core components: a large language model \textbf{(LLM)}, a vision-language model \textbf{(VLM)}, and a visual navigation model \textbf{(VNM)}. The LLM is a generative model based on the transformer architecture, trained on internet text, and it parses textual instructions into a sequence of landmarks. The \textbf{VLM}, specifically the CLIP \cite{radford2021learning}, encodes images and text into an embedding space, allowing it to associate images with textual descriptions. By computing the cosine similarity between the \textbf{VLM} embeddings of landmarks and images, probabilities are obtained to align landmark descriptions with images. The \textbf{VNM} is used to infer the robot's navigation effectiveness between nodes in a graph, based on estimated temporal distances. These components work together by parsing instructions into landmarks, associating landmarks with graph nodes, and optimizing a probabilistic objective to find the optimal path on the graph, which is then executed using the \textbf{VNM} model~\cite{shah2022lmnav}. 

LM-Nav employs a graph search algorithm to determine the optimal sequence of nodes to visit in order to follow the text instructions. For each landmark \(l_i\) extracted by the language model, the vision-language model computes a probability \(P(v|l_i)\) that every node \(v\) in the graph corresponds to \(l_i\). It defines a function \(Q(i,v)\) representing the maximum score for reaching node \(v\) while visiting landmarks up to \(l_i\). \(Q(0,v)\) is initialized based on the shortest path from the start. Then for each landmark \(i=1,...,n\), it updates \(Q(i,v)\) either by carrying over the previous \(Q(i-1,v)\) and adding \(P(v|l_i)\), or by transitioning from a neighboring node \(w\) with \(Q(i,w) - \alpha \cdot D(v,w)\), where \(D(v,w)\) is the travel cost between \(v\) and \(w\). After computing \(Q(n,v)\) for all \(v\), it chooses the node with the maximum value as the destination. The optimal path is recovered by backtracking through the \(Q\) values. However, a potential limitation is that summing \textbf{VLM} probabilities across landmarks could cause the system to visit an incorrect node for landmark \(l_i\) if the cumulative past scores overwhelm the current \(P(v|l_i)\), deviating from the intended instructions.

\subsection{Role of Transformers}

Transformers play a crucial role in LM-Nav. The Large Language Model (LLM) component, powered by transformers, parses free-form instructions into a list of landmarks \cite{shah2022lmnav}. The Vision-Language Model (VLM) component utilizes transformers to associate these landmarks with nodes in a graph by estimating the probability of correspondence. In the Visual Navigation Model (VNM) component, transformers estimate navigational affordances and robot actions, enabling navigation between graph nodes \cite{shah2023vint}. LM-Nav leverages pre-trained models like GPT-3 and CLIP (\textit{ViT-L-14}), which have been trained on large visual and language datasets, to execute complex high-level instructions without requiring user-annotated navigational data. The use of transformers across all components allows LM-Nav to seamlessly integrate language understanding, visual perception, and navigation capabilities, facilitating robust robotic navigation from natural language instructions.

\subsection{Assumed Semantics}

For grounding landmark text with images, LM-Nav uses the CLIP model, specifically the \textit{ViT-L-14} variation of it. The core assumed semantics of the CLIP model is its ability to associate and align visual and textual representations in a shared embedding space. Through contrastive pretraining on a massive dataset of image-text pairs, the model learns to map semantically related images and texts close together in this embedding space, while pushing unrelated pairs apart. This learned alignment between visual and linguistic modalities allows the model to effectively capture and leverage the semantic relationships between images and their corresponding textual descriptions, captions, or labels. By projecting both image and text inputs into a common embedding space, CLIP can understand and reason about the visual-semantic alignments, enabling it to perform tasks that require bridging the visual and textual domains, such as image-text retrieval, image captioning, and multimodal classification.
\section{Representation Vulnerabilities of Visual Language Models}
\subsection{Equivalence Structures of the Embedding Space.} 
In our earlier work~\cite{salman2024intriguing}, we have proposed a framework to explore and analyze the embedding space of vision transformers, revealing intriguing equivalence structures and their implications for model robustness and generalization.  Understanding the structures of the representation space is crucial, as they determine how the model generalizes. In general, we model the representation given by a (deep) neural network (including a transformer) as a function $f: \mathbb{R}^m\rightarrow\mathbb{R}^n$. A fundamental question is to have a computationally efficient and effective way to explore the embeddings of inputs by finding the inputs whose representation will match the one given by $f(x_{tg})$, where $x_{tg}$ is an input whose embedding we like to match. Informally, given an image of a stop sign as an example, all the images that share its representation given by a model will be treated as a stop sign.

\subsection{Embedding Alignment Procedure}
We describe the algorithm for embedding alignment (\cite{salman2024intriguing,salman2024unalign}), aimed at matching the representation of an input (image embedding) to that of a target input (e.g., image or text embedding). To accommodate the requirement of aligning two vectors, a loss function  L(x) is defined to quantify the difference between the embedding of a modified input ($x_0+\Delta x$) and the target embedding ($f(x_{tg})$). The goal is to minimize this loss, which is represented as the squared norm of the difference between these two embeddings. 
\begin{equation}
   L(x) = L(x_0+\Delta x)= \frac{1}{2}\Vert f(x_0+\Delta x) - f(x_{tg})\Vert^2,
\end{equation}
The method involves calculating the gradient of the loss function concerning the input, which is related to the Jacobian of the representation function at the initial input($x_0$). This gradient information is used to adjust the input in a way that minimizes the loss, moving the input's embedding closer to the target's embedding. 

Although obtaining the optimal solutions might involve solving either a quadratic or linear programming problem, depending on the chosen norm for minimizing $\Delta x$, the gradient function has proven to work effectively for all the cases we have tested, attributable to the Jacobian of the transformer. 
Note that automatic gradient calculations are supported by all deep learning frameworks\footnote{We have used PyTorch for our experiments.} and are done efficiently as the required time is the same as one step backpropagation for one sample. 

\section{Adversarial Route Manipulation Method}

In this section, we describe an adversarial route manipulation method designed to redirect a visual landmark-based navigation system from its intended path to a specific malicious target node. The method operates on a graph representation of the environment, where nodes correspond to distinct viewpoints and are associated with one or more images (e.g., front and back views). Our approach alters the graph to deceive the navigation system, which usually plans the best route by matching visual observations to landmark descriptions. This makes the system assume it has visited all landmarks while heading to the target malicious destination.

\begin{algorithm}
\caption{Finding Optimal Nodes for Representation Matching}
\label{alg:optimal_nodes}
\textbf{Input}: Graph $G = (V, E)$, start node $s \in V$, target node $t \in V$, landmark text embeddings $\{l_1, l_2, \ldots, l_n\}$ \\\\
\textbf{Output}: Sequence of nodes $\{v_1, v_2, \ldots, v_{n-1}, t\}$ for representation matching

    \begin{algorithmic}[1]
        \STATE $P \gets $ ShortestPath($G, s, t$) \COMMENT{Find shortest path from start to target node}
        \STATE $m \gets |P|$ \COMMENT{Number of nodes in the shortest path}
        \STATE $\mathbf{S} \gets $ GetSimilarityMatrix($P, \{l_1, l_2, \ldots, l_n\}$) \COMMENT{Compute similarity matrix}
        \STATE $\mathbf{D} \gets \text{initialize\_dp}(m, n)$ \COMMENT{Initialize DP table}
        \STATE $\mathbf{D}[0][0] \gets 0$
        \STATE $\mathbf{P'} \gets \text{initialize\_parent}(m, n)$ \COMMENT{Initialize parent table for backtracking}
        \STATE $\mathbf{D}, \mathbf{P'} \gets \text{dynamic\_programming}(\mathbf{S}, \mathbf{D}, \mathbf{P'})$ \COMMENT{Populate DP table}
        \STATE $\mathbf{l_i}, \mathbf{v_j} \gets \text{backtrack}(\mathbf{P'}, m, n)$ \COMMENT{Backtrack to find optimal landmarks}
        \STATE $\{v_1, v_2, \ldots, v_{n-1}\} \gets \text{extract\_results}(\mathbf{S}, \mathbf{P}, \mathbf{l_i}, \mathbf{v_j})$ \COMMENT{Extract best landmark nodes}
        \STATE $v_n \gets t$ \COMMENT{Include the target node as the final node}
        \STATE \textbf{return} $\{v_1, v_2, \ldots, v_n\}$
    \end{algorithmic}
\end{algorithm}

The proposed method consists of two key algorithms. Algorithm \ref{alg:optimal_nodes} focuses on the optimal selection of nodes for modification along the shortest path between the start and target nodes. We employ Dijkstra's algorithm to determine this path, and then, with Dynamic Programming (DP), we sequentially identify the nodes that exhibit the highest similarity scores with respect to the corresponding landmark text embeddings. If we have $m$ nodes in the path and $n$ landmarks, then the DP algorithm's time complexity will be $\mathcal{O}(m \times n)$. This selection strategy ensures that the chosen nodes are already well-aligned with the landmark representations, facilitating faster convergence and higher similarity scores during the subsequent optimization phase. Importantly, the target node is specified as the final landmark, and the route manipulation algorithm is designed to direct the robot along the malicious trajectory, maximizing its progression toward the designated target node.

\begin{algorithm}
\caption{Adversarial Node Modification}
\label{alg:node_modification}
\textbf{Input}: Graph $G = (V, E)$, optimal node sequence $\{v_1, v_2, \ldots, v_n\}$, landmark text embeddings $\{l_1, l_2, \ldots, l_n\}$ \\
\textbf{Output}: Modified graph $G' = (V', E)$

\begin{algorithmic}[1]
\STATE $V' \gets V$ \COMMENT{Initialize modified node set}

\FOR{$i = 1$ to $n$}
    \STATE $v_i \gets $ ModifyNodeWithTextEmb($v_i, l_i$) \COMMENT{Optimize node w.r.t. landmark text}
    \STATE $N_i \gets \{u \in V' \mid \text{Similarity}(u, l_i) > \text{Similarity}(v_i, l_i)\}$ \COMMENT{Find nodes with higher similarity}
    \STATE $t_i \gets $ FindTargetImage($G, l_i$) \COMMENT{Find target image with lowest similarity}
    \FOR{$u \in N_i$}
        \STATE $u' \gets $ ModifyNodeWithImgEmb($u, t_i$) \COMMENT{Decrease similarity w.r.t. target image}
        \STATE $V' \gets V' \setminus \{u\} \cup \{u'\}$ \COMMENT{Update modified node set}
    \ENDFOR
\ENDFOR

\STATE $G' \gets (V', E)$ \COMMENT{Construct modified graph}
\STATE \textbf{return} $G'$
\end{algorithmic}
\end{algorithm}

Algorithm \ref{alg:node_modification} optimizes the representations of the chosen nodes to increase their similarity with the corresponding landmark text embeddings. It leverages the algorithm introduced in our previous works (\cite{salman2024intriguing,salman2024unalign}), where text embedding is used as the target input, effectively aligning image and text representations. Subsequently, for each optimized landmark node, we identify a set of nodes within the graph that exhibit higher similarity scores compared to that landmark node. We then systematically reduce the similarity of these nodes with respect to a target image embedding from the graph, which matches the corresponding landmark with the lowest similarity. This similarity reduction process also employs the algorithm from our prior work \cite{salman2024intriguing}. Finally, we update the graph with the modified node representations, effectively creating a deceptive route that guides the navigation system toward the target node while maintaining the illusion of visiting all intended landmarks.

The key advantage of our approach lies in its ability to minimize the number of nodes requiring modification to decrease similarity, thereby reducing the overall modification effort. By first increasing the similarity of the selected nodes to the landmark text embeddings, we achieve higher similarity scores, consequently reducing the number of nodes needed for subsequent similarity reduction. This strategic combination of similarity enhancement and reduction steps enables efficient adversarial route manipulation.
\begin{figure*}[ht]
  \centering
  \vspace{-0.20in}
  \includegraphics[width=0.95\textwidth]
  {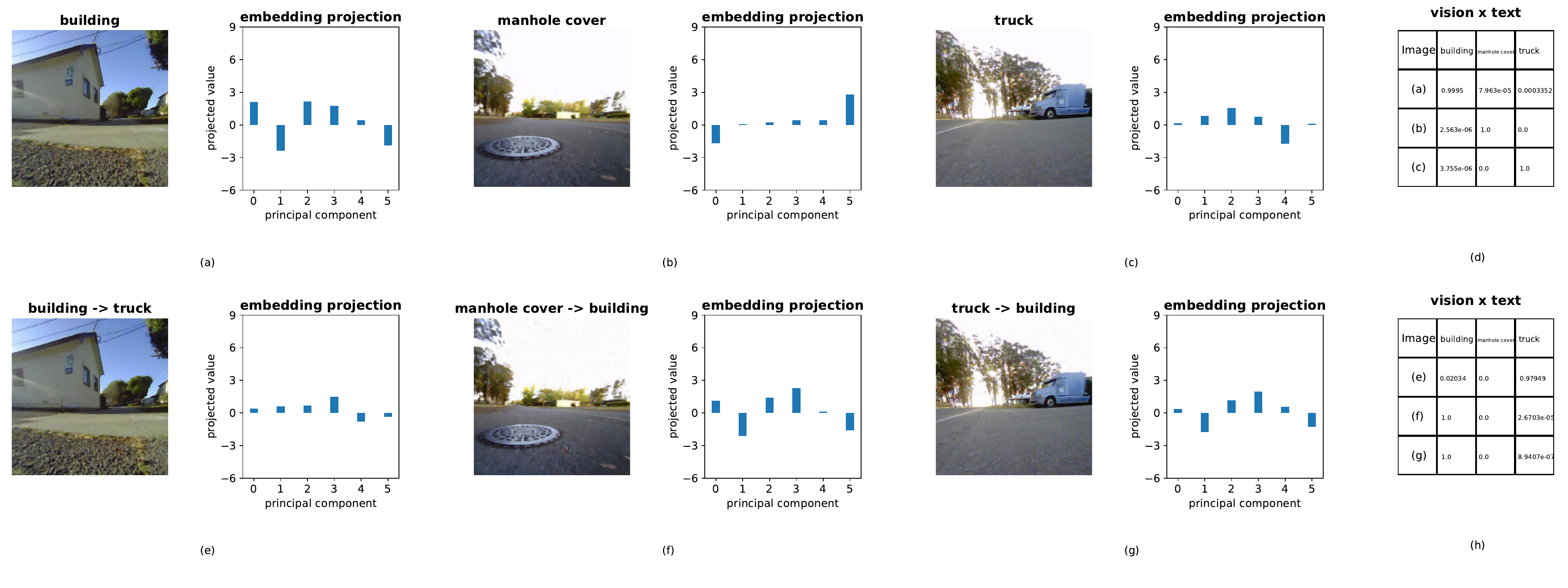}\label{fig:projection_embeddings1}
  \vspace{-0.25in}
  \caption{Typical examples from the RECON dataset obtained using the proposed framework in \cite{salman2024intriguing}. Three pairs of visually indistinguishable images (a and e, b and f, c and g) have different representations from each other as shown in their low-dimensional projections. 
  In contrast, images (f), (g), and (a) exhibit highly similar representations despite their significant semantic differences. A similar trend is observed with images (e) and (c). Please note that the arrow in the title ($original \rightarrow target$) indicates a derived image resulting from aligning the embedding of the original image with that of the target image using the method described. The matrices (d) and (h) show the classification outcomes from the multimodal CLIP pre-trained model used directly with no modifications.
}
  \label{fig:overall_1}
  \vspace{-0.15in}
\end{figure*}

\section{Experimental Results}
\subsection{Dataset and Settings}

To evaluate the efficacy of our method, we utilize two graphs previously employed in the LM-Nav system. These graphs were constructed using the RECON dataset{\footnote{https://sites.google.com/view/recon-robot/dataset}}, which comprises 5000 self-supervised trajectories and observations gathered from a robot navigating through an environment. The graph construction process involves the Visual Navigation Model (VNM), which leverages a goal-conditioned distance function to discern connectivity among raw observations, thereby facilitating the creation of a topological graph. This model constructs the graph by using image and GPS data collected while manually operating the robot. The two graphs differ in size: EnvLarge-10, which contains $278$ nodes, and EnvSmall-10, with $241$ nodes. Each node in these graphs contains two images (capturing both the rear and front views) along with GPS observations and information on connections to adjacent nodes. These images are taken by the robot's onboard camera as it moves through the environment. Our experimentation included 10 distinct paths for each graph, each path featuring a varying number of landmarks. All experiments have been conducted on a laboratory workstation equipped with two NVIDIA A5000 GPUs. 
The code for our proposed method and experiments can be found in this GitHub repository \footnote{https://github.com/programminglove08/RobustnessVisualNav}. 

\subsection{Representation Matching}

As outlined in Algorithm \ref{alg:node_modification}, once we have derived the optimal path with selected nodes, we proceed to modify them based on another target embedding representation. We acquire the input image embedding using CLIP and subsequently generate a target embedding from either a target image or target text, based on the specific requirement. We then calculate the loss function and gradient. We update the input image based on the gradient until it reaches a predetermined L2 distance and cosine similarity for the two representations. For the chosen node, we increase the similarity of the image representation by using the landmark text embedding. In the later part, when decreasing a node's similarity for a specific landmark, we use another target image's embedding, which has the lowest similarity in the graph for that specific landmark. We compare the modified image with the original image using SSIM (Structural Similarity Index) and PSNR (Peak Signal-to-Noise Ratio)~\cite{hore2010ssim}, assessing structural similarity and image quality respectively. SSIM produces a value between 0 and 1, where 1 indicates perfect similarity.  PSNR is expressed in decibels (dB), and higher values indicate better image quality (i.e., less distortion).

In the context of image-to-image representation similarity matching, Figure \ref{fig:overall_1} presents three examples. Images a, b, and c each exhibit distinct images alongside their respective unique embedding projections. However, when image a is modified with image c (resulting in image e), image b with image a (resulting in image f), and image c with image a (resulting in image g), the original images exhibit representation projections that closely resemble those of the target images. In terms of image-to-image representation matching, the average PSNR across all modified nodes in the graph for the original and modified images stands at 42.86 dB. Additionally, the average SSIM score is 0.975.

\begin{figure*}[ht]
  \centering
  \vspace{-0.20in}
  \includegraphics[width=0.95\textwidth]
  {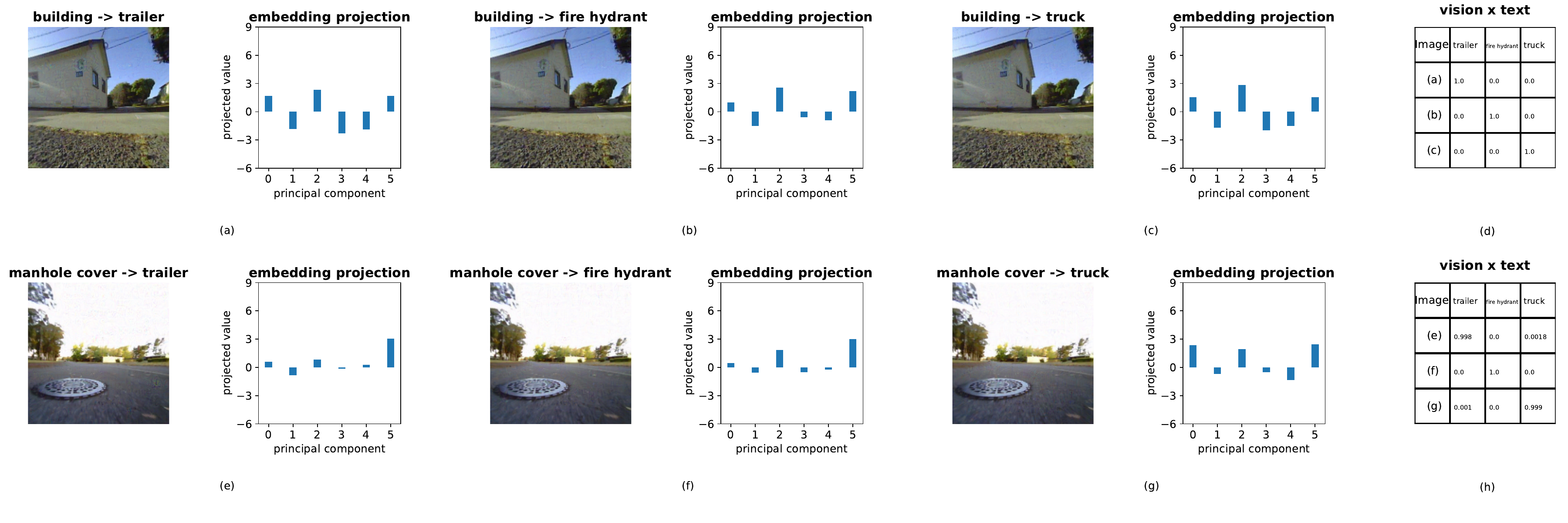}\label{fig:projection_embeddings2}
  \vspace{-0.25in}
  \caption{Additional examples from the RECON dataset obtained using the proposed framework in \cite{salman2024intriguing}; however, the arrow in the title ($original \rightarrow target$) here signifies a derived image from the original one by aligning the embedding of the original image with the target text embedding using the method. The projections of embedding-aligned images closely resemble the projections of the aligned text.
}
  \label{fig:overall_2}
  \vspace{-0.15in}
\end{figure*}

\begin{figure}[ht]
  \centering
\vspace{-0.15in}
   \includegraphics[width=0.47\textwidth]{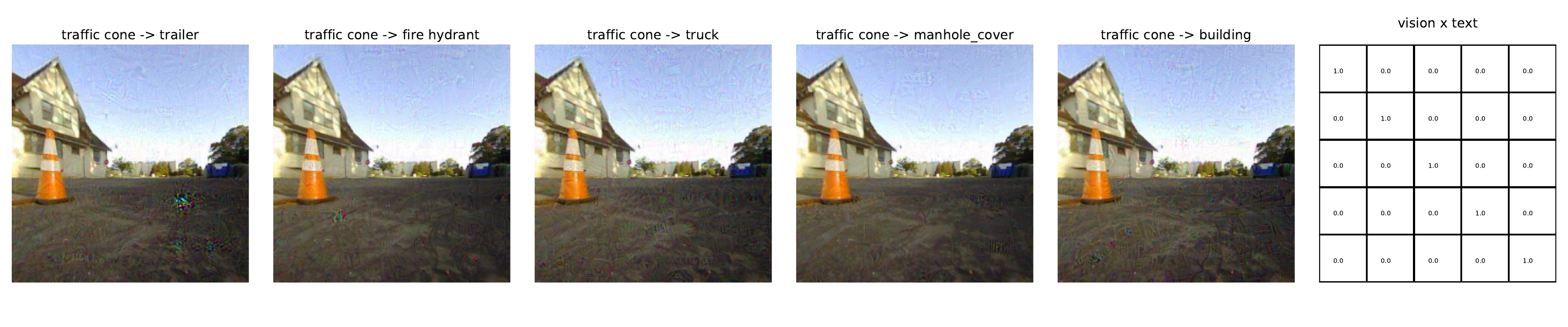}\\

   \includegraphics[width=0.47\textwidth]{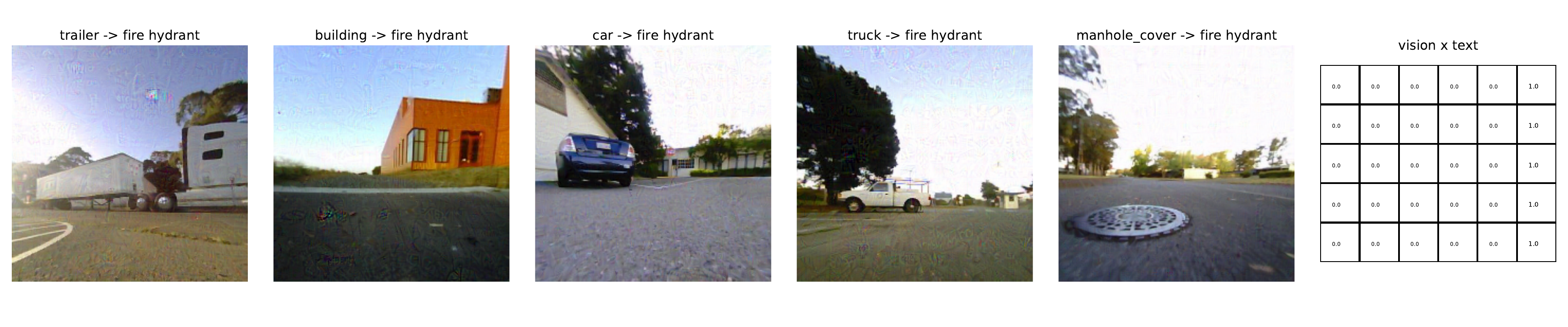}\\
\vspace{-0.15in}
  \caption{
  (top) More instances from the RECON dataset, where visually indistinguishable images have very different representations via embedding alignment with the corresponding texts, and resulting significantly different classification outcomes (as shown in the classification probabilities; each row in the matrix corresponds to one image (from left to right)). (bottom) Visually distinguishable images have very similar embeddings, aligned and classified to a particular text.
}
  \label{fig:more_examples}
  \vspace{-0.15in}
\end{figure}

Figure \ref{fig:overall_2} showcases the effectiveness of image-to-text representation similarity matching, illustrating that two initially dissimilar images with distinct embedding projections can yield similar projections after optimization with the same text. To demonstrate the universal applicability of the procedure and the adversarial examples that exist almost everywhere, Figure \ref{fig:more_examples} shows more examples from different landmarks from the RECON dataset. In this representation matching mode, the average PSNR for both original and modified images across all nodes is 40.45 dB, with an average SSIM score of 0.962.

\begin{figure}[ht]
  \centering
  \includegraphics[width=0.4\textwidth]{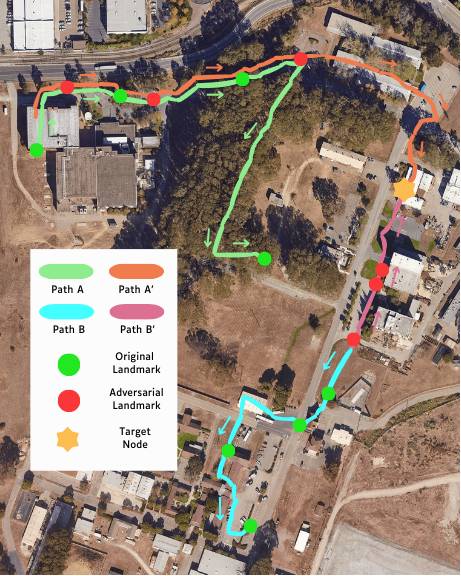}
  \caption{Visualization of robot route manipulation in EnvLarge-10, showing original Paths A (\textbf{Green}) and B (\textbf{Cyan}) with their respective landmarks. The green nodes denote original landmarks. Paths A and B are adversarially modified into Paths $A'$ (\textbf{Orange}) and $B'$ (\textbf{Pink}) with red nodes modified to resemble corresponding landmarks. The LM-Nav system guides the robot along these modified paths to reach the targeted malicious node, denoted by \textbf{Yellow Star}. Here, the first half of Path A and Path A' is shown separately for a clear view.}
  \label{fig:example_route}
\end{figure}

\begin{table}[!htbp]
\centering
\begin{tabular}{|l|l|l|}
\hline
\textbf{Environment} & EnvLarge-10 & EnvSmall-10 \\ \hline
\textbf{Route Modification Success Rate} & 100\% & 100\% \\ \hline 
\textbf{Modified Landmark Matching Rate} & 91.7\% & 79.9\% \\ \hline
\textbf{Path Efficiency} & 98.7\% & 89.2\% \\ \hline
\textbf{Arrival Success Rate} & 90\% & 30\% \\ \hline
\end{tabular}
  \vspace{0.10in}
\caption{Performance Metrics of Robotic Route Manipulation in Large and Small Environments}
\label{tab:my_label}
\end{table}

\subsection{Systematic Evaluation}

Across the two example environments with 20 paths, we successfully manipulate routes by transforming specific nodes into other landmarks or modifying specific nodes to decrease the similarity for a particular landmark. The modified images generate routes consistent with the adversarial landmarks while retaining visual indistinguishability. With these manipulated routes, the robot successfully redirects to different paths while moving toward a malicious target node. For both image-to-text and image-to-image representation matching, our optimization step runs for 12,000 steps. In image-to-text matching, we employ a learning rate of 0.5, an L2 distance threshold of 269, and a cosine similarity threshold of 0.75. Conversely, in image-to-image matching, we utilize a learning rate of 0.09, an L2 distance threshold of 16, and a cosine similarity threshold of 0.95. Figure \ref{fig:example_route} depicts two paths from EnvLarge-10, Path A (containing landmarks: \textit{[`a building with a red-black wall', `a fire hydrant', `a grove', `a manhole cover', `a trailer']}) and Path B (containing landmarks: \textit{[`a white building', `a white truck', `a square with a large tree', `a stop sign']}), which are adversarially modified and transformed into Path $A'$ and Path $B'$, respectively. Utilizing the LM-Nav system and the modified graph, the robot follows the malicious routes and reaches the targeted malicious node. 

We have conducted four distinct experiments, the results of which are summarized in Table \ref{tab:my_label}. In each experiment, we have selected a specific node as the target malicious node in both large and small environments, modifying the graph for each of the 10 paths. Our goal is to assess the robot's success in navigating from the starting node to the target malicious node, traversing through intermediate nodes treated as landmarks, with the target node as the final landmark. We have evaluated the performance based on several metrics:

\textbf{Route Modification Success Rate:} This metric reflects the proportion of instances where the robot is successfully redirected to an alternative route, ultimately reaching a different final node than originally intended. Our results show a 100\% success rate in diverting the robot's route across all 20 paths in both environments.

\textbf{Modified Landmark Matching Rate:} This measures the accuracy with which the robot identifies the nodes along the malicious path as the intended landmarks. In the large environment, the robot accurately identifies the modified node as the intended landmark 91.7\% of the time, whereas in the small environment, this rate is 79.9\%.

\textbf{Path Efficiency:} This metric indicates the percentage overlap between the path proposed by our algorithm and the planned path by the LM-Nav system. We have observed a 98.7\% match in the large environment and an 89.2\% match in the small environment.

\textbf{Arrival Success Rate:} The robot successfully reaches the target malicious node 90\% of the time in the large environment but only 30\% in the small environment. Despite the path efficiency of nearly 90\% in the small environment, the lower arrival success rate is attributed to the LM-Nav system's design, as discussed in section III.A. This design may cause the robot to revisit previous nodes with high similarity scores, deviating from the intended path. While we enhance the final landmark similarity scores across all paths, in some instances within both large and small environments, the optimized nodes of earlier landmarks display high similarity scores. This led to a scenario where the cumulative similarity favors these nodes over the target node, mistakenly guiding the robot to a previous node and resulting in a failed arrival at the final destination.

\section{Robust Detection of Adversarial Image Modifications}
The main idea revolves around the robust detection of adversarial image modifications by leveraging the detectable artifacts induced by such modifications in the feature representations (embeddings) produced by the VLM (in our case CLIP) model. The algorithm compares the feature representations produced by the CLIP model for original and adversarially modified images, both with and without added Gaussian noise~\cite{salman2024zshot}. For batches of images, it computes the average difference in feature representations between the original and noisy versions. The key idea is that within a specific range of noise levels, the feature representation differences will be significantly larger for adversarially modified images compared to original images, enabling accurate detection of such modifications.

    \begin{figure}[ht]
        \centering
        \includegraphics[width=0.4\textwidth]{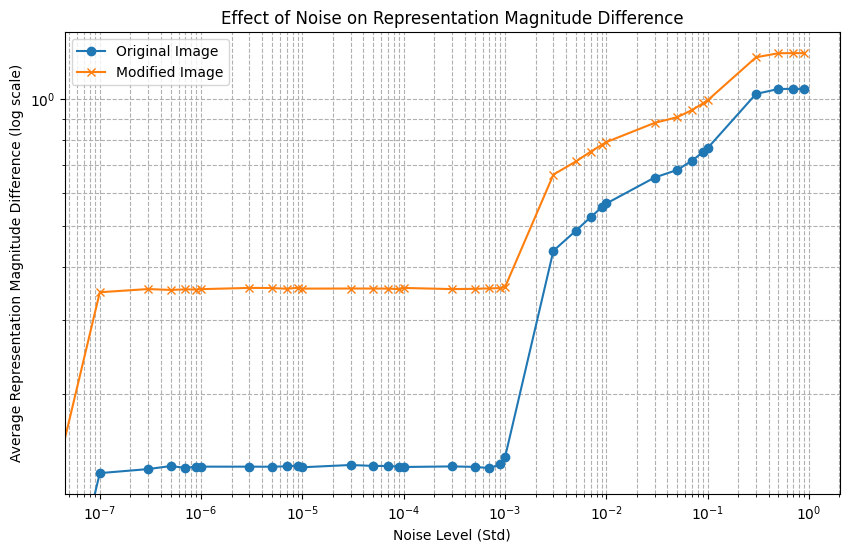}
        \caption{The detection of adversarial image modifications via feature representation (embedding) analysis with noise variation in the CLIP model.}
        \label{fig:adv_detection}
    \end{figure}

For this experiment, we consider a graph with $278$ nodes, each containing two images: a front view and a back view. We then adversarially modify each image within the nodes with the help of target embedding representing a different landmark that does not originally exist in that node. We run the algorithm on both the original and modified graphs, varying the noise standard deviation from $10^{-7}$ to $9\times10^{-1}$. Figure \ref{fig:adv_detection} represents the result of the algorithm. Within the noise level range of $10^{-7}$ to $2\times10^{-3}$, the differences between the feature representations of the original and modified images are significant, enabling accurate differentiation. Therefore, employing a noise level with a standard deviation of 
$10^{-5}$ represents an optimal choice for detecting these modifications, as it efficiently highlights the critical differences we aim to identify. At a noise standard deviation level of 
$10^{-5}$ and using a feature map difference threshold of $0.203$, the detection task achieved an accuracy of $0.963$ and an F1 score of $0.961$ for a balanced sample set of $278$ nodes. This suggests that noise levels within this range could be effective for detecting adversarial modifications by analyzing the changes in the feature representations produced by the CLIP model. Consequently, for a given path containing multiple nodes, if the detection algorithm identifies certain nodes as being more sensitive than others, we can reliably infer that this path might have been modified through an adversarial attack. However, it is important to note that this effective range may vary across different datasets.

\section{Discussions}
In this paper, we have demonstrated a fully working method that can modify all the routes in the LM-Nav system~\cite{shah2022lmnav} and tested it on their datasets. While our route manipulation algorithm is designed specifically for the system, fundamentally our embedding matching algorithm is model agnostic and is effective on all existing vision transformer-based models (\cite{salman2024intriguing,salman2024unalign}) and therefore the route manipulation algorithm can be changed to exploit the representation vulnerabilities in the vision transformers a VLN system relies on.

While the LM-Nav system enables a robot to navigate complex outdoor environments, the noise and large variations in lighting and other conditions can render the representations of images less reliable. As the system uses the sum of the cosine similarities
between an image and all the landmarks, the navigation algorithm may not always work as expected. 
The lower final node arrival success rate in Tab. \ref{tab:my_label} than the 100\% route modification rate, is due to this unexpected property of their routing algorithm. Our algorithm can also be used to
increase the sharpness of landmarks, resulting in more accurate routes. 

In this paper, we focus on stealth modifications to images stored in databases; since imperceptible changes to the images can cause the representations to change significantly, such changes can not be identified visually. 
Note that our algorithm can also be modified to design specific physical objects and patterns that can be added to the scene to modify the representations and therefore change the 
robot's routes maliciously. 
While such physical 
objects and patterns have been demonstrated successfully to adversarially attack deep-learning-based sensors for autonomous driving (\cite{inproceedings,10.1145/3460120.3485377}), 
we anticipate more reliable and robust behaviors 
using our designed patterns
as our algorithm can match the underlying representations of chosen images and
the feasibility of such objects to exploit representation vulnerabilities is being explored.

The ability to stealthily control robot behavior via adversarial graph modifications highlights inherent vulnerabilities in today's visual language navigation systems. More broadly, as vision-language models are being tested for autonomous driving, the representation vulnerabilities of such models (such as GPT-4V~\cite{wen2023road})  must be tested thoroughly.


\section {Conclusion and Future Work}
In this work, we have demonstrated the first method to adversarially attack vision-language robot navigation systems by imperceptibly modifying the images, which result in significant changes to their representations and cause the navigation algorithms to produce very different routes.
Further progress requires building semantics and robustness into learned representations even though the proposed change detection of adversarial modifications is an effective first solution. 

More broadly, as artificial intelligence techniques are built into more robotic and autonomous systems, studying their inherent and fundamental vulnerabilities becomes more critical. 
As these systems can affect people's lives and cause significant damage if exploited, their vulnerabilities must be tested thoroughly; in many ways, AI techniques have become what we rely on and 
vulnerabilities in infrastructures and 
operating systems provide the most relevant and valuable lessons~\cite{Gorbenko_OS_Vulnerability_17}.
While some robust training methods could improve the robustness of chosen attacks for classifiers~\cite{Muhammad_22_Adversarial}, how they improve the inherent representation vulnerabilities 
of vision transformers have not been studied systematically yet, which is being investigated.

\bibliographystyle{IEEEtran}
\bibliography{reference}

\end{document}